\documentclass[letterpaper]{article} % DO NOT CHANGE THIS
\usepackage{aaai2026} 
\nocopyright

\usepackage{tabularx}
\usepackage{booktabs}
\usepackage{makecell}
\usepackage{amsmath}
\usepackage{algorithmic}
\usepackage{multirow}
\usepackage{multicol}
\usepackage{cite}
\usepackage{amsfonts}
\usepackage{times}  % DO NOT CHANGE THIS
\usepackage{helvet}  % DO NOT CHANGE THIS
\usepackage{courier}  % DO NOT CHANGE THIS
\usepackage[hyphens]{url}  % DO NOT CHANGE THIS
\usepackage{graphicx} % DO NOT CHANGE THIS
\usepackage{natbib}  % DO NOT CHANGE THIS AND DO NOT ADD ANY OPTIONS TO IT
\usepackage{caption} % DO NOT CHANGE THIS AND DO NOT ADD ANY OPTIONS TO IT
\usepackage{algorithm}
\usepackage{algorithmic}

\usepackage{newfloat}
\usepackage{listings}
\DeclareCaptionStyle{ruled}{labelfont=normalfont,labelsep=colon,strut=off} % DO NOT CHANGE THIS
\floatstyle{ruled}
\newfloat{listing}{tb}{lst}{}
\floatname{listing}{Listing}
\title{Uncertainty-Aware Spatial Color Correlation for Low-Light Image Enhancement}
\author {
    Jin Kuang\textsuperscript{\rm 1,2,3},
    Dong Liu\textsuperscript{\rm 1,2}\footnote{corresponding author.},
    Yukuang Zhang\textsuperscript{\rm 4},
    Shengsheng Wang\textsuperscript{\rm 4}
}
\affiliations {
    \textsuperscript{\rm 1}School of Computer and Artificial Intelligence, Xiangnan University\\
    \textsuperscript{\rm 2}Hunan Engineering Research Center of Advanced Embedded Computing and Intelligent Medical Systems, Xiangnan University\\
    \textsuperscript{\rm 3}School of Geosciences Yangtze University\\
    \textsuperscript{\rm 4}College of Computer Science and Technology, Jilin University \\ 
    gasking.stu@yangtzeu.edu.com, liudong@xnu.edu.com, zyk24@mails.jlu.edu.com, wss@jlu.edu.cn
}

\begin{document}

\maketitle

\begin{abstract}
Most existing low-light image enhancement approaches primarily focus on architectural innovations, while often overlooking the intrinsic uncertainty within feature representations particularly under extremely dark conditions where degraded gradient and noise dominance severely impair model reliability and causal reasoning. To address these issues, we propose U$^2$CLLIE, a novel framework that integrates uncertainty-aware enhancement and spatial-color causal correlation modeling. From the perspective of entropy-based uncertainty, our framework introduces two key components: (1) An Uncertainty-Aware Dual-domain Denoise (\textit{\textit{UaD}}) Module, which leverages Gaussian-Guided Adaptive Frequency Domain Feature Enhancement (\textit{\textit{G2AF}}) to suppress frequency-domain noise and optimize entropy-driven representations. This module enhances spatial texture extraction and frequency-domain noise suppression/structure refinement, effectively mitigating gradient vanishing and noise dominance. (2) A hierarchical causality-aware framework, where a Luminance Enhancement Network (\textit{LEN}) first performs coarse brightness enhancement on dark regions. Then, during the encoder-decoder phase, two asymmetric causal correlation modeling modules—Neighborhood Correlation State Space (\textit{\textit{NeCo}}) and Adaptive Spatial-Color Calibration (\textit{\textit{AsC}})—collaboratively construct hierarchical causal constraints. These modules reconstruct and reinforce neighborhood structure and color consistency in the feature space. Extensive experiments demonstrate that U$^2$CLLIE achieves state-of-the-art performance across multiple benchmark datasets, exhibiting robust performance and strong generalization across various scenes.
\end{abstract}


\section{Introduction}
Low-light image enhancement remains challenging due to noise, color distortion, and detail loss from poor illumination, impairing both visual perception and high-level tasks like detection and nighttime driving. \par
Early methods like histogram equalization, gamma correction, and Retinex-based curve fitting struggle to jointly restore global illumination and preserve local details, often introducing artifacts and noise. Deep learning methods\cite{guo2023low}\cite{wang2022uformer} improve performance by learning global semantics and local features, but still face limitations: (1) CNNs, constrained by local receptive fields, 
\begin{figure}[H]
    \centering
    \includegraphics[width=1.\linewidth]{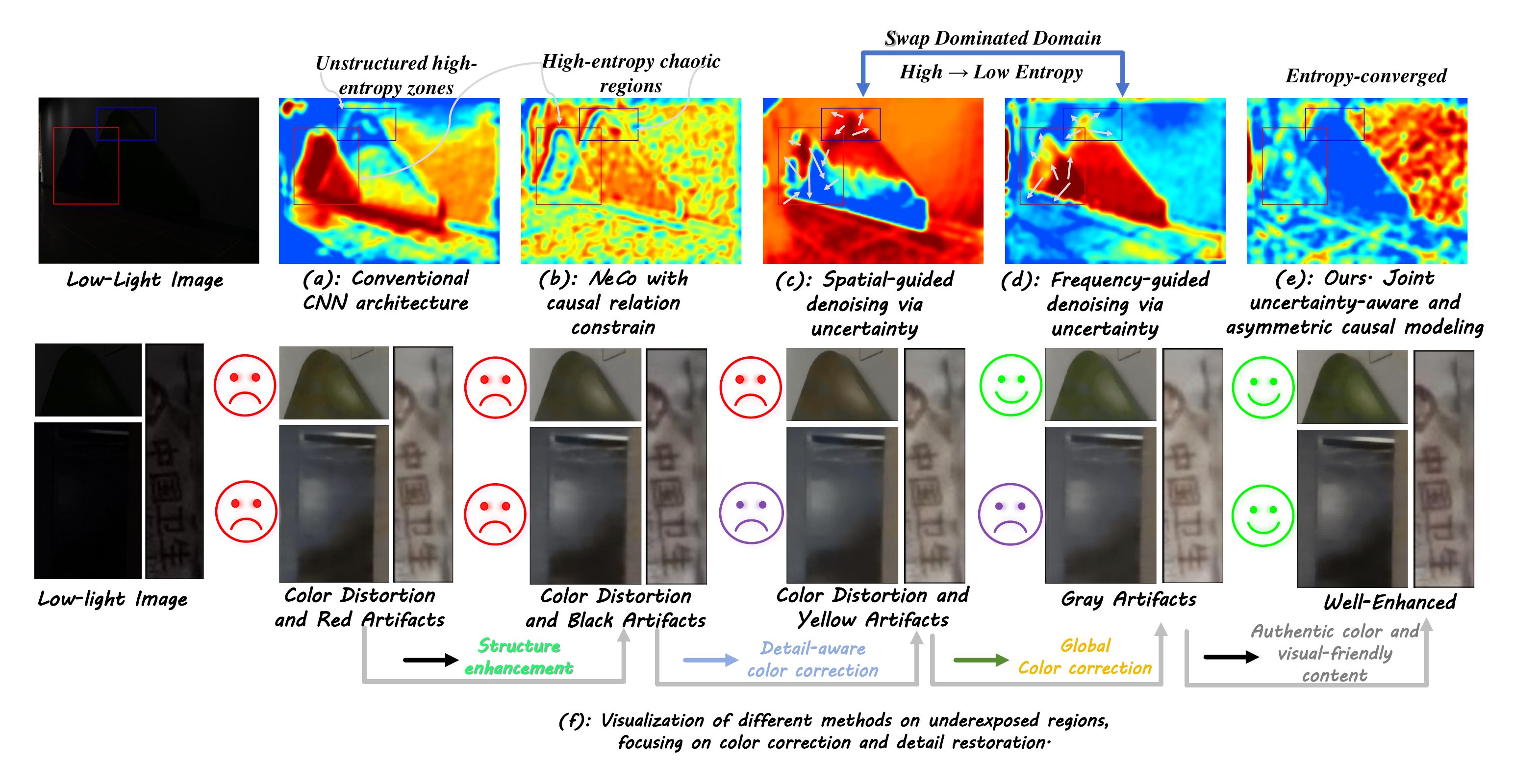}
    \caption{\textbf{Top row} compares entropy maps across five methods: (a) CNN baseline shows disordered entropy; (b) adding Mamba-based causal modeling reduces artifacts; (c) spatial-domain guidance enhances structural consistency; (d) frequency-domain cues improve color and structure fidelity; (e) our method yields the most coherent entropy and fine detail near edges. \textbf{Bottom row} shows corresponding enhancements, where the last column achieves artifact-free, locally consistent color and structure, highlighting the synergy of dual-domain uncertainty guidance and causal modeling.}
    \label{motivation}
\end{figure}
fail to stably model underexposed regions and details, leading to chaotic features and color distortions (see Figure~\ref{motivation}(a):

Red box shows unstable feature entropy causing representation chaos; blue box indicates blurred boundaries); (2) Transformers expand receptive fields via self-attention but suffer from high computational cost. Recent Mamba-based models\cite{gu2023mamba}\cite{liu2024vmamba} offer a balance between global modeling and efficiency via state space models (SSMs). Yet, directly applying Mamba to low-light enhancement is suboptimal due to weak local correlation modeling. Improvements like MambaLLIE\cite{weng2024mamballie} and BSMamba\cite{zhang2025bsmamba} introduce local bias or structure enhancement, but still lack hierarchical consistency across stages, resulting in artifacts (\textit{e.g.}, Figure~\ref{motivation}(f): 3rd column). Another key challenge lies in effective feature guidance. Some methods exploit frequency-aware fusion (FourLLIE\cite{wang2023fourllie}, DMFourLLIE\cite{zhang2024dmfourllie}), semantic constraints (SCL-LLIE\cite{liang2022semantically}), or causal color modeling (SKF\cite{wu2023learning}, SMGLLIE\cite{xu2023low}), while others use SNR-aware pixel enhancement\cite{xu2022snr}. However, most overlook the role of uncertainty (\textit{e.g.}, entropy) in low-light features. As shown in Figure~\ref{motivation}(c–e), feature optimization can be seen as an entropy-reducing process, where lower entropy in dark regions leads to more stable features and better reconstruction (see Figure~\ref{motivation}(f): last column).\par

To tackle these challenges, we propose U$^2$CLLIE (Uncertainty-Aware Spatial Color Correlation for Low-Light Image Enhancement), as shown in Figure~\ref{overframe}. It addresses two core issues: \textbf{(1) Feature guidance collapse} under extreme darkness, where noise overwhelms gradients and disrupts causal color/detail correlation; \textbf{(2) Local perception modeling}, \textit{i.e.}, effectively leveraging sparse low-light cues without strong priors to achieve consistent color and detail reconstruction for perceptual quality restoration.\par

Specifically, U$^2$CLLIE first employs a Luminance Enhancement Network (\textit{LEN}) for global brightness adjustment. A Neighborhood Correlation State Space Module (\textit{\textit{NeCo}}) in the encoder models causal dependencies via learnable neighborhood-aware weights, while an Adaptive Spatial-Color Calibration Module (\textit{\textit{AsC}}) in the decoder captures object-level color correlations. These asymmetric modules impose progressive causal constraints to enhance detail and color reconstruction. To suppress structural noise and recover texture, an Uncertainty-Aware Dual-domain Denoising Module (\textit{\textit{UaD}}) is introduced, which guides enhancement via dual-domain (frequency/spatial) modeling under an entropy minimization perspective. Specifically, the Gaussian-Guided Adaptive Frequency Domain Feature Enhancement (\textit{\textit{G2AF}}) module performs adaptive frequency-domain denoising with Gaussian kernels. Our main contributions are: 

% \noindent 1) Reformulating low-light enhancement as an uncertainty entropy minimization problem, with a \textit{\textit{UaD}} module enabling dual-domain guidance for noise suppression and detail recovery. \par

% \noindent 2) Designing an asymmetric causal modeling strategy via \textit{\textit{NeCo}} (feature-level correlation) and \textit{\textit{AsC}} (color-level correlation), enforcing hierarchical consistency. \par

% \noindent 3) Achieving state-of-the-art performance and strong generalization across benchmark datasets and downstream tasks. \par

\begin{enumerate}
    \item [1) ]Reformulating low-light enhancement as an uncertainty entropy minimization problem, with a \textit{\textit{UaD}} module enabling dual-domain guidance for noise suppression and detail recovery.
    \item [2) ]Designing an asymmetric causal modeling strategy via \textit{\textit{NeCo}} (feature-level correlation) and \textit{\textit{AsC}} (pixel-level correlation), enforcing hierarchical consistency.
    \item [3) ]Achieving state-of-the-art performance and strong generalization across benchmark datasets and downstream tasks.
\end{enumerate}

% \begin{itemize}
%     \item [1) ]Reformulating low-light enhancement as an uncertainty entropy minimization problem, with a \textit{\textit{UaD}} module enabling dual-domain guidance for noise suppression and detail recovery.
%     \item [2) ]Designing an asymmetric causal modeling strategy via \textit{\textit{NeCo}} (feature-level correlation) and \textit{\textit{AsC}} (color-level correlation), enforcing hierarchical consistency.
%     \item [3) ]Achieving state-of-the-art performance and strong generalization across benchmark datasets and downstream tasks.
% \end{itemize}

\section{Related work}
\subsection{Low-Light Image Enhancement}
\textbf{Traditional methods. }Histogram equalization\cite{pisano1998contrast}, gamma correction\cite{wang2009real} directly enhance brightness but ignore the interplay between luminance, noise, and details, often resulting in artifacts and limited effectiveness in real-world scenarios.\\
\textbf{Retinex-based methods. }\cite{wang2013naturalness}\cite{zhang2021beyond} decompose images into illumination and reflection to jointly model brightness and detail. While methods like RetinexFormer\cite{cai2023retinexformer}, RetinexMamba\cite{bai2024retinexmamba}, and Diff-Retinex++\cite{yi2025diff} improve local/global modeling through semantic priors, SSMs, or diffusion models, their performance is limited by the illumination estimation network and increased computational cost, with insufficient exploration of stable causal dependencies.

\begin{figure}[H]
    \centering
    \includegraphics[width=1.\linewidth]{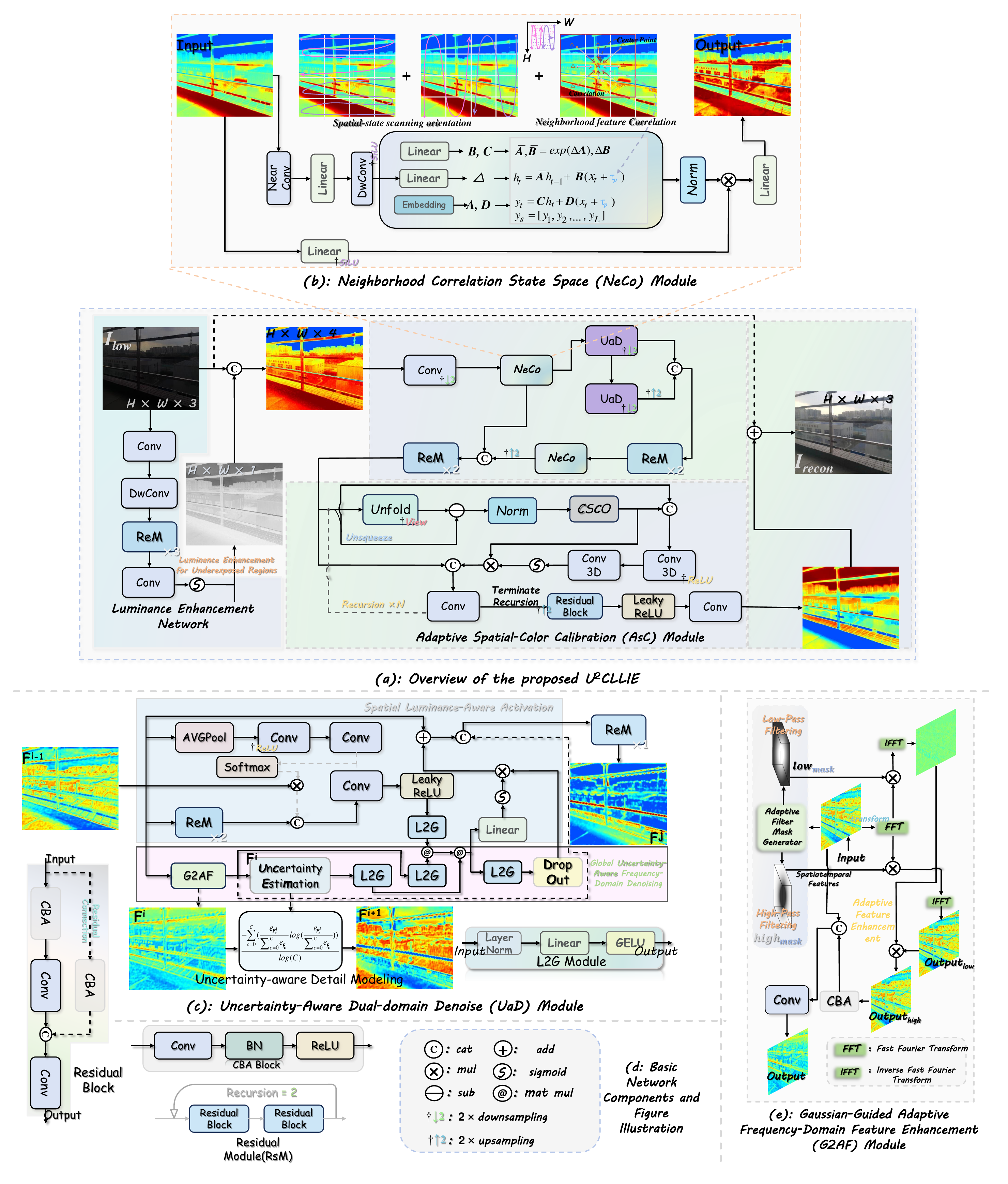}
    \caption{The proposed U$^2$CLLIE. (a) Overview of U$^2$CLLIE, where \textit{LEN} enhances dark regions and suppresses overexposure; \textit{AsC} models color-wise causal correlations with low complexity. (b) \textit{NeCo} captures causal consistency via multi-directional neighborhood scanning. (c) \textit{UaD} refines textures (spatial branch) and suppresses noise (frequency branch) under uncertainty guidance. (d) Basic network structure. (e) \textit{G2AF} adaptively enhances frequency features using Gaussian masks.~(\textit{Please zoom in to the best view.})}
    \label{overframe}
\end{figure}

\textbf{Deep learning-based methods. }\cite{jiang2021enlightengan}\cite{wang2024zero}\cite{yang2023implicit}\cite{liang2022learning}\cite{wang2024correlation}\cite{yan2025hvi}\cite{zhang2025cwnet} leverage CNNs for enhanced local-global feature modeling. To address CNNs’ limited receptive fields, Transformer-based approaches\cite{chen2021pre}\cite{zamir2022restormer} use local window attention for long-range interaction. Frequency-domain models (\textit{e.g.}, FourLLIE, DMFourLLIE) improve structural consistency by enhancing amplitude features. More recently, SSM-based methods (\textit{e.g.}, Wave-Mamba\cite{zou2024wave}, BSMamba, MambaLLIE) offer a balance between efficiency and modeling capacity, yet still struggle with causal consistency in extremely dark regions and adjacent structures.

\subsection{Uncertainty-Aware Entropy-Guided Modulation}
Uncertainty entropy reflects model confidence and feature stability, and has shown strong representational power in tasks like semi-supervised learning\cite{springenberg2015unsupervised}, clustering\cite{jain2017subic}, and domain adaptation\cite{hu2025beyond}\cite{lee2024entropy}. High-entropy regions often correspond to noisy, low-SNR areas with unstable features, while low-entropy regions are typically more reconstructable. In low-light enhancement, entropy has been used to guide learning paradigms—from spatial regression to confidence-aware optimization\cite{zhou2003learning} to improve perceptual quality and reduce noise. However, most methods rely on uncertainty cues in a single domain, limiting their ability to capture spatial and frequency dependencies—particularly in high-entropy regions prone to texture loss and color shifts.

\section{Methodology}

% To address key challenges in low-light image enhancement, we propose U$^2$CLLIE (Uncertainty-Aware Spatial Color Correlation for Low-Light Image Enhancement), targeting two core problems: (1) building robust representations in severely degraded, underexposed regions, and (2) modeling spatial and color-wise causal relationships between neighboring pixels to preserve structural and color consistency. U$^2$CLLIE consists of three main components: a Luminance Enhancement Network (LEN), an encoder–decoder backbone, and several uncertainty- and causality-driven modules. Section 3.1 outlines the overall architecture, while Sections 3.2–3.4 detail each module.

\subsection{Motivation and Overview of the U$^2$CLLIE}
Most low-light enhancement methods adopt residual regression between low-light inputs and normal-light targets. However, for severely underexposed regions, this paradigm suffers from two core problems: 

\noindent \textbf{Problem 1: Feature degradation in dark regions.} Including gradients degradation (\textbf{\textit{Limitation 1}}) and noise dominated (\textbf{\textit{Limitation 2}}).\\
\textbf{Problem 2: Breakdown of neighborhood context modeling.~} \textbf{\textit{See the supplement for formal analysis of these issues.}}\\
Our key motivation is to avoid relying on raw noisy inputs and instead reconstruct robust features via uncertainty-guided and causally-constrained modeling. Accordingly, as shown in Figure~\ref{overframe}, U$^2$CLLIE adopts a U-shaped architecture with following core designs:\\
\textbf{1. Uncertainty-aware entropy guidance (addresses Problem 1):}
\begin{itemize}
    \item \textit{LEN} enhances luminance and provides contextual cues for underexposed regions;

    \item \textit{\textit{UaD}} reconstructs denoised, modelable features under dual-domain (spatial/frequency) guidance; \textit{G2AF} adaptively suppresses frequency-domain noise.
\end{itemize}

\noindent \textbf{2. Hierarchical causal modeling (addresses Problem 2):}
\begin{itemize}
    \item \textit{NeCo} (encoder) captures directional neighborhood causality;
    \item \textit{AsC} (decoder) models structured color-wise correlations for consistent reconstruction.
\end{itemize}

\noindent These components collaboratively enhance SNR, suppress noise/artifacts, and recover structural and color consistency in extremely dark scenes(see Figure~\ref{motivation}(f): last column).

\subsection{Luminance Enhancement Network~(\textit{LEN})}
As the front-end module, \textit{LEN} enhances dark regions to facilitate subsequent feature modeling. It first projects the input  $I_{\text{low}} \in \mathbb{R}^{H \times W \times 3}$ into feature space $\tilde{F}_{\text{ill}} \in \mathbb{R}^{H \times W \times c_1}$  via a $1 \times 1 $ convolution ($c_{1}$=32), followed by a $3 \times 3$ depthwise separable convolution for local detail extraction, and three residual blocks (\textit{ReM}). A Sigmoid activation then produces a spatial luminance-aware map $\hat{F}_{ill} \in \mathbb{R}^{H \times W \times 1}$:

\vspace{-8pt}
\begin{equation}
\begin{aligned}
 \tilde{F}_{\text{ill}} &= Conv(I_{\text{low}}), \\
 \hat{F}_{\text{ill}} &= \sigma\Big(\text{Conv}\big(\text{ReLU}(\text{ReM}_{\times 3}(\text{DepthConv}(\tilde{F}_{\text{ill}})))\big)\Big). 
\end{aligned}
\end{equation}

\textit{LEN} is trained to predict the luminance difference $\hat{I}_{diff}=\frac{I^{ycrcb}_{high}-I^{ycrcb}_{low}}{I^{ycrcb}_{high}}$ between low-light and normal images in the \textit{Y} channel of YCrCb space (ablation comparisons in HSV, Lab, and HVI will be presented in experiment). Its loss($\mathcal{L}_{len}$) is $\frac{1}{B \cdot H \cdot W}\sum_{n=1}^{B}\sum_{i,j}^{H,W}(\hat{I}_{diff}^{(n,i,j)} - \hat{F}_{\text{ill}}^{(n,i,j)})^2.$

% \begin{equation}
%     \mathcal{L}_{len} =\frac{1}{B \cdot H \cdot W}\sum_{n=1}^{B}\sum_{i,j}^{H,W}(\tilde{I}_{diff}^{(n,i,j)} - \hat{F}_{ill}^{(n,i,j)})^2.
% \end{equation}

\subsection{Uncertainty Estimation}
% Existing dual-branch (spatial + frequency) methods face key limitations: (1) A single fusion weight fails to balance global features, biasing optimization toward bright regions and neglecting complex textures in dark areas; (2) Under extreme low light, the structure may degrade into a single path, causing texture–noise confusion and leading to artifacts or overexposure. \par

\noindent \textbf{Uncertainty-Aware Dual-domain Denoise (\textit{\textit{UaD}}) Module.} \textit{\textit{UaD}}  employs entropy-guided suppression of high-uncertainty regions to reduce noise, while enhancing low-entropy, high-frequency details for robust feature extraction in extreme darkness. Given input features $F^{i-1} \in \mathbb{R}^{\frac{H}{2} \times \frac{W}{2} \times c_2}$, the spatial branch generates luminance-sensitive features $F_{\text{spa}}$, while the frequency branch produces entropy-guided features $F^i$. Specifically, $F^{i-1}$ is first enhanced by the \textit{\textit{G2AF}} module, followed by uncertainty entropy computation, yielding $F^{i+1}$. The features undergoes cross-attention and nonlinear to yield $F_{\text{fre}}$. Subsequently, $F^{i-1}$ and $F_{\text{fre}}$ are added pixel-wise($\oplus$) and concatenated with the denoised feature $F^i$, resulting in $F^j \in \mathbb{R}^{\frac{H}{2} \times \frac{H}{2} \times c_2}$. This facilitates multi-domain collaboration to suppress noise and artifacts in high-entropy uncertain regions. 
\begin{equation}
\begin{aligned}\label{entropy:computatio}
\vec{F} &= \text{Softmax}(\text{Conv}^2(\text{AvgPool}(F^{i-1}))) \otimes F^{i-1},\\ 
F_{\text{spa}} &= \text{LeakyReLU}(\text{Conv}(\text{cat}(\vec{F}, \text{ReM}(F^{i-1})))),\\
\mathcal{H}(x) &= -\tfrac{1}{\log c}\textstyle\sum_{c=1}^C p_c(x)\log p_c(x), \\
F^i &= \text{\textit{G2AF}}(F^{i-1}),F^{i+1} = \mathcal{H}(F^i), \\
&\text{Attn}(Q,K,V) = \text{Softmax}(QK^\top\!/\!\sqrt{d_{\text{head}}})V, \\
F_u &= \text{Attn}(F_{\text{spa}},F^i,F^{i+1}),\\ 
F_{\text{fre}} &= \sigma(\text{Linear}(F_u)) \otimes \text{Dropout}(\text{L2G}(F_u)), \\
F^j &= \text{ReM}(\text{cat}(F^{i-1} \oplus F_{\text{fre}}, F^i)). 
\end{aligned}
\end{equation}

Here, $Softmax$, ($\sigma: Sigmoid$), and $LeakyReLU$ serve as a weight normalization function, an independent distribution activation, and a nonlinear activation functions, respectively; $AvgPool$ denotes average pooling; $dim_{head}$ is the shared channel dimension of \textit{Q}, \textit{K}, and \textit{V}; $cat$ indicates concatenation along dimension dim;  $DropOut$ is stochastic neuron dropout; $Conv^i$ denotes convolution applied $i$ times; \par

\noindent \textbf{Gaussian-Guided Adaptive Frequency-Domain Feature Enhancement (\textit{\textit{G2AF}}) Module.} We propose the \textit{G2AF} module, which jointly regulates amplitude and phase while adaptively generating Gaussian mask radii based on frequency feature distributions. Specifically, input feature $F^{i-1}$ is transformed via \textit{FFT} yielding $\tilde{F}^{i-1}_{\text{fre}}$, followed by the \textit{Adaptive Filter Mask Generator} (see Eq.~\ref{g2af} and \textit{pseudocode in the supplementary}) to produce frequency-aware Gaussian masks for low- and high-frequency components ($low_{\text{mask}},high_{\text{mask}} \in \mathbb{R}^{\frac{H}{2} \times \frac{W}{2} \times 1}$). This process can be described as follows:

% \begin{align}
%     dist_x &= linspace(-1,1,\text{H}).\text{view}(1,\text{H},1), \\ \nonumber
%     dist_y &= linspace(-1,1,\text{W}).\text{view}(1,1,\text{W}), \\ \nonumber

%     dist_{mat} &= \sqrt{dist_x^2 + dist_y^2}, \\ \nonumber
    
%     r_{low} &= r_{low} \times Sigmoid(\tilde{F}_{fre}^{i-1}), \\ \nonumber
%     r_{high} &= r_{high} \times Sigmoid(\tilde{F}_{fre}^{i-1}), \\ \nonumber

%     low_{mask} &=exp^{\frac{dist_{mat}^2}{2r_{low}^2 + \epsilon}}, \\ \nonumber
%     high_{mask} &=exp^{\frac{dist_{mat}^2}{2r_{high}^2 + \epsilon}}
% \end{align}

\vspace{-10pt}
\begin{equation}
\begin{aligned}\label{g2af}
\text{dist}_x &= \text{linspace}(-1,1,H).\text{view}(1,H,1), \\
\text{dist}_y &= \text{linspace}(-1,1,W).\text{view}(1,1,W), \\ 
\text{dist} &= \sqrt{\text{dist}_x^2 + \text{dist}_y^2}, \\ 
r_{\text{low}} &\gets r_{\text{low}} \cdot \sigma(\tilde{F}^{i-1}_{\text{fre}}), r_{\text{high}} \gets r_{\text{high}} \cdot \sigma(\tilde{F}^{i-1}_{\text{fre}}), \\
low_{\text{mask}} &= \exp\left( -\text{dist}^2 / (2 r_{\text{low}}^2 + \epsilon) \right), \\ 
high_{\text{mask}} &= \exp\left( -\text{dist}^2 / (2 r_{\text{high}}^2 + \epsilon) \right). 
\end{aligned}
\end{equation}

Subsequently, $F^{i-1}$ and $low_{mask},high_{mask}$ are multiplied pixel-wise and then transformed the spatial domain via \textit{iFFT} to obtain 
$\tilde{F}_{low}^{i-1}$ and $\tilde{F}_{high}^{i-1}$. Then, $\tilde{F}_{low}^{i-1}, \tilde{F}_{high}^{i-1}$ are multiplied($\otimes$) and convolved for local frequency aggregation. The result is concatenated with input feature $F^{i-1}$ and refined via convolution to produce the final feature $F^i \in \mathbb{R}^{\frac{H}{2} \times \frac{W}{2} \times c}$.
\begin{equation}
\begin{aligned}\label{fftcat}
\tilde{F}_{\text{low}}^{i-1} &= \lambda \cdot \text{iFFT}(F^{i-1} \otimes low_{\text{mask}}), \\ 
\tilde{F}_{\text{high}}^{i-1} &= (1-\lambda) \cdot \text{iFFT}(F^{i-1} \otimes high_{\text{mask}}), \\
F^i &= \text{Conv}\big(\text{cat}(\text{CBA}(\tilde{F}_{\text{low}}^{i-1} \oplus \tilde{F}_{\text{high}}^{i-1}), F^{i-1})\big).
\end{aligned}
\end{equation}

In Equations~(\ref{g2af}–\ref{fftcat}), $linspace(\cdot, \cdot, \cdot)$ generates evenly spaced values; $view(\cdot,\cdot, \cdot)$ reshapes feature dimensions; $\lambda$ is a hyperparameter (set to 0.5); $r_{low},r_{high}$ are learnable parameters initialized to 0.3 and 0.1; $\epsilon$ is set to 10$^{-6}$.\newline

\noindent \textbf{How \textit{UaD} Addresses Problem 1:}
\begin{itemize}
    \item \textbf{Overcoming Gradient Vanishing (Limitation 1): }Through the cross-attention mechanism, the \textit{V} term is associated with the uncertainty entropy $F^{i+1}$, leveraging the entropy distribution to address the gradient vanishing issue in dark regions. Specifically, in extremely dark areas (high entropy: $F^{i+1} \rightarrow 1$), the modulation factor ($\hat{I}_{\text{recon}} - I_{\text{high}}$) effectively prevents gradient vanishing:
    \begin{equation}\label{gradient:vanish}
        \frac{\partial \mathcal{L}_{\text{recon}}(x)}{\partial \theta} \propto F^{i+1}\cdot (\hat{I}_{\text{recon}} - I_{\text{high}}) \cdot \frac{\partial F_{\text{fre}}}{\partial \theta}.
    \end{equation}

    \item \textbf{Mitigating Noise Dominance (Limitation 2): }In noise-concentrated regions, the following operation: $F_{\text{fre}} = \phi(\mathcal{H}(F^i))$(where $\phi(\cdot)$ denotes a network operation, and $\mathcal{H}(\cdot)$ is defined in Eq.~\ref{entropy:computatio}).

\vspace{-10pt}
\begin{equation}
\begin{aligned}
&\frac{\partial F_{\text{fre}}}{\partial \theta} \propto \frac{\partial \phi(\mathcal{H}(F^i))}{\partial \mathcal{H}} \cdot \frac{\partial\mathcal{H}(F^i)}{\partial F^i} \cdot \frac{\partial F^i}{\partial \theta} \\
&\Rightarrow \frac{\partial \mathcal{L}_{\text{recon}}(x)}{\partial \theta} \propto F^{i+1} \cdot (\hat{I}_{\text{recon}} - I_{\text{high}}) \cdot \frac{\partial F_{\text{fre}}}{\partial \theta},\\
&\Rightarrow \underbrace{F^{i+1}}_{\in (0,1)} \cdot \underbrace{(\hat{I}_{\text{recon}} - I_{\text{high}})}_{> 0} \cdot \underbrace{\frac{\partial \phi(\mathcal{H}(F^i))}{\partial \mathcal{H}} \cdot \frac{\partial \mathcal{H}(F^i)}{\partial F^i} \cdot \frac{\partial F^i}{\partial \theta}}_{\neq 0} \\
&\Rightarrow
\mathcal{E}_{\text{noise}} \leq \gamma \cdot \left\| \frac{\partial F^i}{\partial \theta} \right\|_{Upper~ Bound},~\text{with } \gamma \in (0,1) \label{eq:noise_bound}
\end{aligned}
\end{equation}
\end{itemize}
%%%%%%%%%%%%%%%%%%%%%%%
%%%%%%%%%%%%%%%%%%%%%%
%%%%%%%%%%%%%%%%%%%%%%

% \begin{equation}
%     \frac{\partial F_{fre}}{\partial \theta} \propto \frac{\partial \phi(\mathcal{H}(F^i))}{\partial F^i} \cdot \frac{\partial F^i}{\partial \theta}
% \end{equation}

% \begin{eqnarray}

%      & \rightarrow &\frac{\partial \mathcal{L}_{recon}(x)}{\partial \theta} \propto F^{i+1} \cdot (\hat{I}_{recon} - I_{high}) \cdot \frac{\partial F_{fre}}{\partial \theta}, \\ \nonumber

%      &\rightarrow &\frac{\partial \mathcal{L}_{recon}(x)}{\partial \theta} \propto \underbrace{F^{i+1}}{\in (0,1)} \cdot \underbrace{(\hat{I}_{recon} - I_{high})}{\in (0,1)} \frac{\partial \phi(\mathcal{H}(F^i)}{\partial F^i} \cdot \frac{F^i}{\partial \theta}, \\ \nonumber

%     &\rightarrow &\|N(p)\frac{\partial f_{\theta}}{\partial \theta\]}\|_{noise} \geq \|\gamma \codt \frac{\partial F^i}{\partial \theta}\|_{uncertainty}
% \end{eqnarray} 
% \end{itemize}

% \noindent In Eq.~\ref{eq:noise_bound}, the left side reflects the network fitting noise when $I_{high}(p)$ is hard to optimize. Please refer to \textit{\textbf{the supplementary material}} for detailed derivations and symbol definitions. The uncertainty entropy steers parameter updates toward easier directions, thus mitigating noise fitting in challenging low-light regions. 

\noindent In Eq.~\ref{eq:noise_bound}, the left side reflects the network fitting noise, while the right side demonstrates that uncertainty entropy guides the noise toward a likelihood upper bound. The uncertainty entropy steers parameter updates toward easier directions, thus mitigating noise fitting in challenging low-light regions. \textit{See supplementary for derivations and symbol definitions}.

\subsection{Causal Relationship-Aware Context Modeling}
U$^2$CLLIE introduces an asymmetric encoder-decoder: the encoder models neighborhood causality (\textit{\textit{NeCo}}), and the decoder enhances local color consistency (\textit{\textit{AsC}}), effectively constraining pixel-wise color features via adjacent causal cues to resolve Problem 2.\par

\noindent \textbf{Neighborhood Correlation State Space (\textit{\textit{NeCo}}) Module.} 
Neighborhood convolution is first applied to input feature $F_1 \in \mathbb{R}^{H \times W \times c}$ to extract local context. Meanwhile, during multi-directional scanning, a learnable offset ($\tau_p \in \mathbb{R}^{4 \times dim}$) enhances SSM local states via adjacent causal cues. This forms the basis of our \textit{\textit{NeCo}} module (see Figure~\ref{overframe}):

\vspace{-12pt}
\begin{equation}
    h_t = \bar{\textbf{A}}h_{t-1}+\bar{\textbf{B}}(x_t+\tau_p),y_t=\textbf{C}h_t+\textbf{D}(x_t+\tau_p).
\end{equation}
Here, $\tau_p$ encodes local causal correlations with spatial states; $\bar{\textbf{A}},\bar{\textbf{B}},\textbf{C},\textbf{D}$ are detailed in Figure~\ref{overframe}(b).

The model maintains local consistency during multi-directional scanning while suppressing redundant correlations. Input $F_1 \in \mathbb{R}^{H \times W \times c}$ is processed in two branches: one extracts local constraints using neighborhood convolution, projection, depthwise convolution, and \textit{SiLU} activation, followed by fine-scale convolutions with learnable offsets; the other applies a linear layer and \textit{SiLU}. The outputs are multiplied and projected back to the original dimension:

\vspace{-8pt}
\begin{equation}
\begin{aligned}\label{ssm}
\tilde{F}_1 = &\text{Norm}(\text{2DSSM}(\text{SiLU}( \\
    &\text{DwConv}(\text{Linear}(\text{NearConv}(F_1)))))), \\
\tilde{F}_2 = &\text{SiLU}(\text{Linear}(F_1)), \tilde{F}_{\text{out}} = \text{Linear}(\tilde{F}_1 \otimes \tilde{F}_2).
\end{aligned}    
\end{equation}

Here, \textit{Norm} is LayerNorm layer; \textit{SiLU} is the activation function; \textit{DwConv} is the depthwise separable convolution; \textit{Linear} is the linear projection; \textit{2DSSM} is our SSM-based scanner(see Eq.~\ref{ssm}); \par

% \begin{figure}[htbp]
%     \centering
%     \includegraphics[width=1.\linewidth]{Figure3.pdf}
%     \caption{Color \textit{feature} Space \textit{Cor}relation \textit{Opt}imization. (Explicit Causal Modeling of Central-Distance Correlations)}
%     \label{fig:enter-label}
% \end{figure}

\noindent \textbf{Adaptive Spatial-Color Calibration (\textit{\textit{AsC}}) Module. }
Efficient neighborhood causal modeling is vital in decoding for capturing color consistency. The proposed \textit{\textit{AsC}} module operates as follows: Given feature map $\hat{F}$, we extract local patches via \textit{Unfold} to obtain $\hat{F}_{\text{patch}}$, and map $\hat{F}$ to a central feature  $\hat{F}_c$. After normalizing distances, the top-\textit{k} (in this work, \textit{k}=8) nearest neighbors are selected to form $\hat{F}_{\text{neighbor}}$. Concatenating $\hat{F}_c$ and $\hat{F}_{\text{neighbor}}$, a 3D convolution and Sigmoid yield correlation weights $\hat{F}_{\text{cluster}}$. Then, $\hat{F}_{\text{cluster}},\hat{F}_{\text{neighbor}}$ are element-wise multiplied and summed along the feature-distance dimension to obtain $\hat{F}_{\text{relation}}$. To mitigate sparsity, $\hat{F}_{\text{relation}}$ is fused with $\hat{F}$ via a $1 \times 1$ convolution, yielding the final output $\hat{F}_{\textit{\text{asc}}} \in \mathbb{R}^{H \times W \times c}$ with enhanced color-aware causal modeling:

% \begin{align*}
%     \hat{F}_{patch}=view(unfold(\hat{F})), \hat{F}_c = unsqueeze(\hat{F}), \\ \nonumber

%     \hat{F}_{neighbor} =CscO(Norm(\hat{F}_{patch} - \hat{F}_{neighbor})), \\  \nonumber

%     \hat{F}_{cluster} = Sigmoid(Conv3D^2(cat((\hat{F}_c,\hat{F}_{meighbor}),dim=-1)))
% \end{align*}
\vspace{-12pt}
\begin{equation}
\begin{aligned}
\hat{F}_{\text{patch}} &= \text{view}(\text{Unfold}(\hat{F})), 
\hat{F}_c = \text{unsqueeze}(\hat{F}), \\
\hat{F}_{\text{neighbor}} &= \text{CscO}(\text{Norm}(\hat{F}_{\text{patch}} - \hat{F}_{\text{neighbor}})), \\
\hat{F}_{\text{cluster}} &= \sigma(\text{Conv3D}^2(\text{cat}((\hat{F}_c, \hat{F}_{\text{neighbor}})))), \\
\hat{F}_{\text{relation}} &= \sum_{c=0}^C(\hat{F}_{\text{cluster}}^{c} \cdot \hat{F}_{\text{neighbor}}^{c}), \\
\hat{F}_{\text{asc}} &= Conv(cat(\hat{F}_{\text{relation}}, \hat{F})).
\end{aligned}    
\end{equation}

Here, \textit{view(·)} reshapes features; \textit{unsqueeze} adds a dimension; $Conv3D^i$ denotes the \textit{i}-th 3D convolution operation. \textit{CscO} \textit{denotes Color feature Space Correlation Optimization (further details are provided in the supplementary material)}.\newline

\noindent \textbf{How \textit{NeCo} combined with \textit{AsC} addresses Problem 2: } The asymmetric collaboration between \textit{NeCo} and \textit{AsC} effectively resolves the breakdown of neighborhood context modeling in low-light images. (1) \textit{NeCo} introduces learnable causal weights to model brightness and texture dependencies within local regions, enhancing feature-level neighborhood constraints. (2) \textit{AsC} adaptively computes similarity within feature blocks to model object-level color correlations, ensuring high intra-object color consistency. (3) \textit{NeCo} guides \textit{AsC}’s alignment, while \textit{AsC} reinforces \textit{NeCo}’s structural modeling. This closed-loop feedback enhances the capture of stable causal dependencies under noise and weak signals. (see Figure~\ref{motivation}(f)).\par

\setcounter{table}{1}
\begin{table*}[!t]
  \centering
  \setlength{\tabcolsep}{1mm}
   
    \begin{tabular}{c|c|ccc|ccc|cc}
    
    \toprule
    \multirow{2}[2]{*}{Method} & \multirow{2}[2]{*}{Venue} & \multicolumn{3}{c|}{LOLv2-Real} & \multicolumn{3}{c|}{LOLv2-Synthetic} & \multicolumn{2}{c}{Complexity~(M/G)} \\
          &       & PSNR↑ & SSIM↑ & LPIPS↓ & PSNR↑ & SSIM↑ & LPIPS↓ & \makecell[c]{Params} & \makecell[c]{FLOPs} \\
    \midrule
    NPE\cite{wang2013naturalness}  & TIP 2013 & 17.33  & 0.464 & 0.2359 & 16.60  & 0.778 & 0.1079 & -     & - \\
    LIME\cite{guo2016lime}  & TIP 2016 & 15.24  & 0.419  & 0.2203 & 16.88 & 0.758 & 0.1041 & -     & - \\
    % RetinexNet & BMVC 2018 & 15.47  & 0.567  &       & 17.13 & 0.798 &       & 0.84  & 587.47 \\
    % DeepUPE & CVPR 2019 & 13.27  & 0.452  &       & 15.08 & 0.623 &       & 1.02  & 21.10  \\
    MIRNet\cite{zamir2020learning} & ECCV 2020 & 20.02  & 0.820  &  -     & 21.94 & 0.876 &    -   & 31.76 & 785.00  \\
    % DRBN  & TIP 2021 & 20.29  & 0.831 & -     & 23.22 & 0.927 & -     & 5.27  & - \\
    SGM\cite{yang2021sparse}   & TIP 2021 & 20.06  & 0.816  & \underline{0.0727} & 22.05 & 0.909 & 0.4841 & 2.31  & - \\
    EnlightenGAN\cite{jiang2021enlightengan} & TIP 2021 & 18.23  & 0.617 & 0.3090  & 16.57 & 0.734 & 0.2200  & 114.35 & 61.01 \\
    IPT\cite{chen2021pre}   & CVPR 2021 & 19.80  & 0.813 & -     & 18.30  & 0.811 & -     & 115.31 & - \\
    FECNet\cite{huang2022deep} & ECCV 2022 & 20.67  & 0.795  & 0.0995 & 22.57 & 0.894 & 0.0699 & 0.15  & 5.82 \\
    SNR-Aware\cite{xu2022snr} & CVPR 2022  & 21.48  & \underline{0.848}  & 0.0740  & 24.13 & 0.927 & 0.0318 & 39.12 & 26.35 \\
    % Uformer & CVPR 2022 & 18.82 & 0.771 & -     & 19.66 & 0.871 & -     & 5.29  & - \\
    Restormer\cite{zamir2022restormer} & CVPR 2022 & 19.94 & 0.827 & -     & 21.41 & 0.830  & -     & 26.13 & - \\
    Bread\cite{guo2023low} & IJCV 2023 & 20.83 & {0.847 } & 0.1740  & 17.63 & 0.919  & 0.0910 & 2.02  & 19.85 \\
    % PairLIE & CVPR 2023 & 19.89  & 0.778 & 0.3170  & 19.07 & 0.794 & 0.2300  & 0.33  & 20.81 \\
    SNR-SKF\cite{wu2023learning} & CVPR 2023 & 20.66 & 0.812 & 0.0757 & 17.21 & 0.774 & 0.0731 & 39.44 & 27.88 \\
    CLE Diffusion\cite{yin2023cle} & MM 2023 & \underline{21.99} & 0.798 & 0.1910  & 20.28 & 0.852 & 0.1020  & 37.02 & 3102.41 \\
    UHDFour\cite{li2023embedding} & ICLR 2023 & 19.42 & 0.790 & 0.1151 & 23.64 & 0.900  & 0.0341 & 17.54 & 4.78 \\
    FourLLIE\cite{wang2023fourllie} & MM 2023 & 22.34 & 0.847  & \textbf{0.0573} & 24.65 & 0.919 & 0.0389 & 0.12  & 4.07 \\
    GSAD\cite{hou2023global}  & NeurIPS 2024 & 20.15  & {0.846}  & 0.1130  & 24.47 & \underline{0.929} & 0.0510  & 17.36 & 442.02 \\
    % Q\textit{UaD}Prior & CVPR 2024 & 20.48 & 0.811 & 0.2020  & 16.11 & 0.758 & 0.1140  & 1252.75 & 1103.20  \\
    DiffUIR\cite{zheng2024selective} & CVPR 2024 & 19.71 & 0.825 & 0.2110  & 19.61 & 0.863 & 0.1600  & 36.26  & 9.88 \\
    UHDFormer\cite{wang2024correlation} & AAAI 2024 & 19.71 & 0.832 & 0.0758  & \underline{24.48} & 0.928 & \underline{0.0310}  & 0.34  & 3.24 \\
    Ours  & ---   & \textbf{22.63} & \textbf{0.851}  & 0.1005 & \textbf{25.15} & \textbf{0.931} & \textbf{0.0308} & 0.48  & 6.18 \\
    
    \bottomrule
    \end{tabular}%
    \caption{Quantitative comparison on LOLv2-Real, LOLv2-Synthetic. The \textbf{best} and \underline{second-best} results are shown in \textbf{bold} and \underline{underline} respectively. \textit{Please note that we did not use the GT-Mean strategy.}}
  \label{tab_real_syn}%
\end{table*}%

\setcounter{figure}{3}
\begin{figure*}[!t]
    \centering
    \includegraphics[width=1.\linewidth]{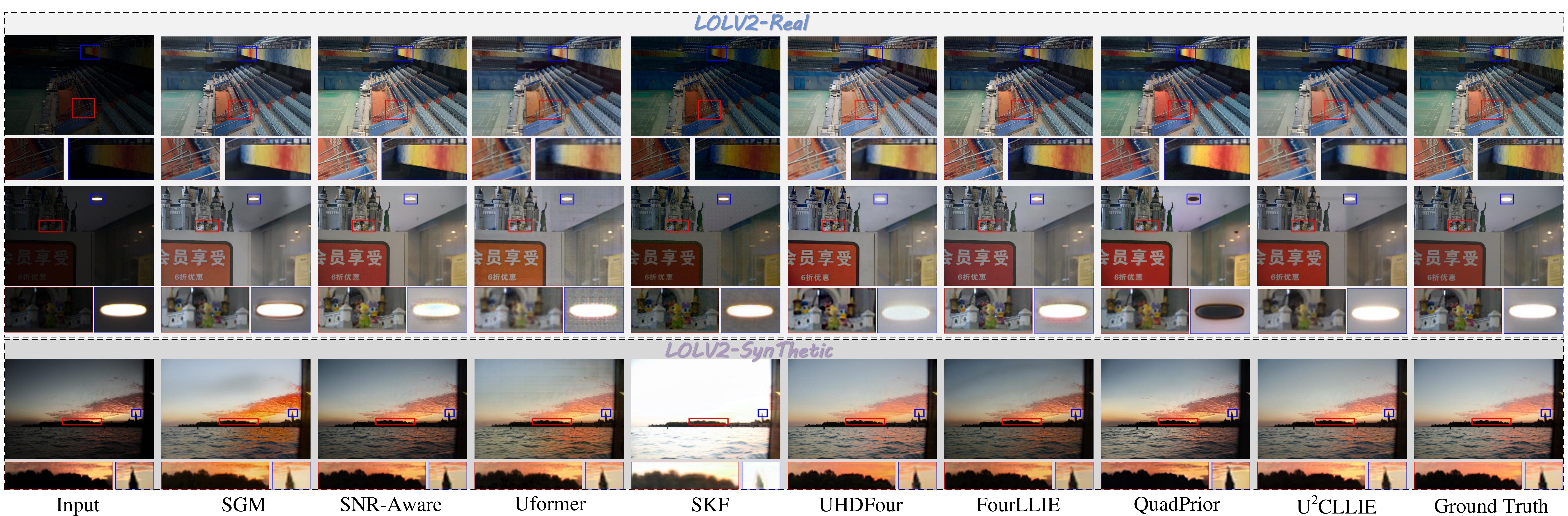}
    \caption{Qualitative comparisons with state-of-the-art methods on the LOLv2-Real and LOLv2-Synthetic datasets demonstrate the superiority of our approach. Our U$^2$CLLIE restores local details more accurately, suppresses artifacts and noise effectively, and preserves consistent color transitions.(\textit{Please zoom in to the best view.})}
    \label{vis_real_synthetic}
\end{figure*}

% \noindent feature blocks to model object-level color correlations, ensuring high intra-object color consistency. (3) \textit{NeCo} provides semantic structure for \textit{AsC}’s color alignment, while \textit{AsC}, in turn, reinforces and validates \textit{NeCo}’s neighborhood modeling. This closed-loop feedback significantly improves the model’s ability to capture stable causal dependencies under noise and weak signals (see Figure~\ref{motivation}(f)).\par

\subsection{Loss Function}
The total loss function of U$^2$CLLIE consists of seven terms:

\vspace{-15pt}
\begin{align}
    \mathcal{L}_{recon} = 
    &\lambda_1\mathcal{L}_{mse} + 
    \lambda_2\mathcal{L}_{ssim} + 
    \lambda_3\mathcal{L}_{per} + 
    \lambda_4\mathcal{L}_{global} + \notag \\
    &\lambda_5\mathcal{L}_{color} + 
    \lambda_6\mathcal{L}_{grad} + 
    \lambda_7\mathcal{L}_{len}.
\end{align}

Where, $\lambda_{1:7} = [0.95,0.01,0.01,0.1,0.5,0.1,0.1]$. $\mathcal{L}_{mse}$: L2 pixel-wise reconstruction loss; $\mathcal{L}_{ssim}$: Structural similarity loss; $\mathcal{L}_{per}$: Perceptual loss (L1 on VGG features); $\mathcal{L}_{global}$: Global histogram alignment via \textit{KL divergence} after \textit{Softmax}; $\mathcal{L}_{color}$: Color consistency (cosine similarity on normalized $\hat{I}_{recon}$/$I_{high}$); $\mathcal{L}_{grad}$: Boundary detail loss (L1 on high-frequency gradients); $\mathcal{L}_{len}$: Lighting difference loss from the \textit{LEN} module. 
Derivations for $\mathcal{L}_{global}$, $\mathcal{L}_{color}$, and $\mathcal{L}_{grad}$ are in the \textbf{supplementary material}.

\section{Experiment}
\subsection{Benchmark Datasets and Settings}
\noindent \textbf{Datasets.} We evaluate our method on 8 publicly available benchmark datasets: 
\noindent \textbf{\textit{1) Reference datasets:} } LOLv2-Real\cite{yang2021sparse} (689 real-world training/100 testing pairs); LOLv2-Synthetic\cite{yang2021sparse} (900 synthetic training/100 testing pairs); LSRW-Huawei\cite{hai2023r2rnet} (2,450 training/30 testing pairs, Huawei P40 captured); \textbf{\textit{2) Unpaired datasets:} }DICM\cite{lee2013contrast} (69 images), LIME (10 images), MEF\cite{ma2015perceptual} (17 images), NPE\cite{wang2013naturalness} (85 images), and VV(24 images) are used to evaluate perceptual quality in the absence of ground-truth references.

% \begin{itemize}
%     \item [\textbf{1)}] \textbf{Reference datasets: }LOLv2-Real (689 real-world training/100 testing pairs), (2) LOLv2-Synthetic (900 synthetic training/100 testing pairs), and (3) LSRW-Huawei (3,150 training/20 testing pairs, Huawei P40 captured).

%     \item [\textbf{2)}] \textbf{Unpaired datasets: }DICM (64 images), LIME (10 images), MEF (17 images), NPE (84 images), and VV (24 images) are used to evaluate perceptual quality in the absence of ground-truth references.
% \end{itemize}

\noindent \textbf{Implementation Details. }All experiments are conducted on an Intel 6326 CPU with a NVIDIA A100 GPU under Ubuntu 20.04.4 LTS. We use PyTorch 1.12.1 with CUDA 12.4. Input images are randomly cropped to 384×384 and augmented via random horizontal and vertical flipping. The batch size is 4. We use the Adam optimizer ($\beta_1$=0.95, $\beta_2$=0.99) with an initial learning rate of 2.5×10$^{-4}$, adjusted via a multi-step scheduler. The model is trained for 1.5×10$^5$ iterations. \par

\noindent \textbf{Evaluation Metrics.} For reference datasets, we report PSNR, SSIM\cite{wang2004image}, and LPIPS (with AlexNet backbone).~For unpaired datasets, perceptual quality is assessed using NIQE\cite{mittal2012making}.

% \begin{figure}[H]
%     \centering
%     \includegraphics[width=1.\linewidth]{Figure5.pdf}
%     \caption{Qualitative results on LRSW-Huawei show that U2CLLIE achieves superior global brightness and local structure preservation, effectively avoiding oversmoothing. ~(\textit{Please zoom in to the best view.})
% }
%     \label{vis_huawei}
% \end{figure}

\subsection{Main Results on Benchmarks}
\noindent \textbf{Quantitative Comparison. }As shown in Tables~\ref{tab_huawei} and~\ref{tab_real_syn}, U$^2$CLLIE achieves the highest PSNR and SSIM across all three reference datasets, and the best LPIPS on LOLv2-Synthetic. Our entropy-guided feature-level approach outperforms not only the pixel-level SNR-based SNR-Aware,

\setcounter{table}{0}
\setcounter{figure}{2}
\begin{table}[H]
  \centering
 \setlength{\tabcolsep}{1mm}
  \begin{tabular}{l|ccc|cc}
    \toprule
    \makecell[c]{Method} & 
    \makecell{PSNR$\uparrow$} & 
    \makecell{SSIM$\uparrow$} & 
    \makecell{LPIPS$\downarrow$} & 
    \makecell{\#Params\\(M)} & 
    \makecell{\#FLOPs\\(G)} \\
    \midrule
    \makecell[c]{NPE} & 17.08 & 0.391 & 0.230 & -    & - \\
    \makecell[c]{LIME} & 17.00 & 0.382 & 0.207 & -    & - \\
    \makecell[c]{Kind} & 16.56 & 0.569 & 0.226 & 8.02  & 34.99 \\
    \makecell[c]{MIRNet} & 19.98 & 0.609 & 0.215 & 31.76 & 785.00 \\
    \makecell[c]{SGM} & 15.43 & 0.570 & 0.237 & 2.31  & - \\
    \makecell[c]{FECNet} & 21.09 & 0.612 & 0.234 & 0.15  & 5.82 \\
    \makecell[c]{HDMNet} & 20.81 & 0.607 & 0.238 & 2.32  & - \\
    \makecell[c]{SNR-Aware} & 20.67 & 0.591 & 0.192 & 39.12 & 26.35 \\
    \makecell[c]{FourLLIE} & 21.11 & 0.626 & 0.183 & 0.12  & 4.07 \\ \midrule
    \makecell[c]{UHDFour} & 19.39 & 0.601 & 0.247 & 17.54 & 4.78 \\ \midrule
    \makecell[c]{Retinexformer} & \underline{21.23} & \underline{0.631} & 0.170 & 1.61  & 15.57 \\ \midrule
    \makecell[c]{UHDFormer} & 20.64 & 0.624 & 0.181 & 0.34  & 3.24 \\ \midrule
    \makecell[c]{RetinexMamba} & 20.88 & 0.630& 0.169 & 3.59  & 34.76 \\ \midrule
    \makecell[c]{CIDNet} & 20.47 & 0.613 & \textbf{0.138} & 1.88  & 7.57 \\ \midrule
    \makecell[c]{Ours} & \textbf{21.35} & \textbf{0.633} & \underline{0.162} & 0.48  & 6.18 \\
    \bottomrule
  \end{tabular}
 \caption{Quantitative comparison on LSRW-Huawei. The \textbf{best} and \underline{second-best} results are shown in \textbf{bold} and \underline{underline} respectively. \textit{Note: no GT-Mean strategy was used.}}
  \label{tab_huawei}
\end{table}

\begin{figure}[H]
    \centering
    \includegraphics[width=1.\linewidth]{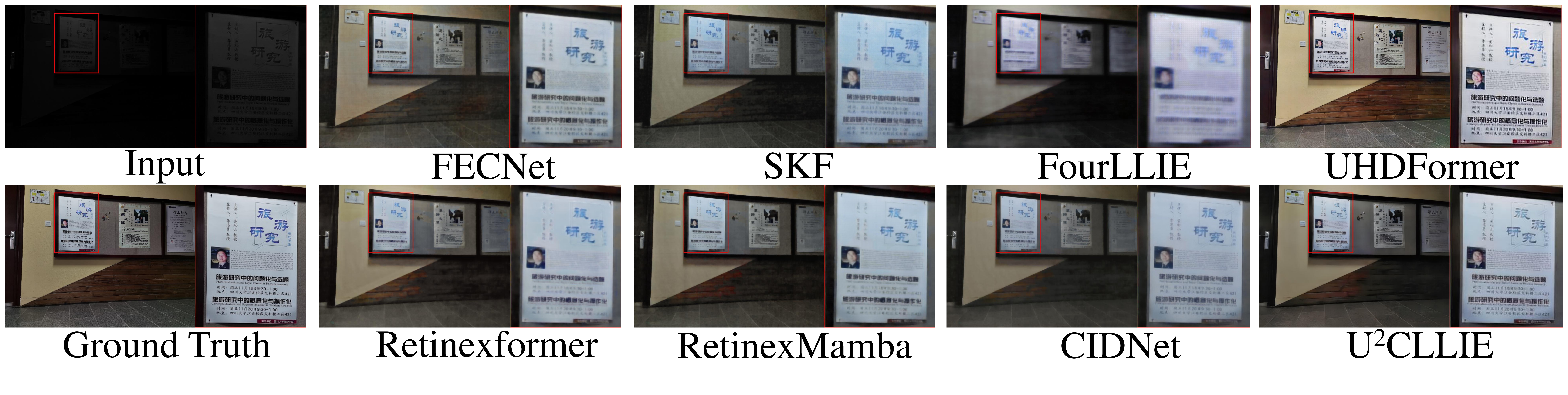}
    \caption{Qualitative results on LSRW-Huawei show that U2CLLIE achieves superior global brightness and local structure preservation, effectively avoiding oversmoothing. ~(\textit{Please zoom in to the best view.})
}
    \label{vis_huawei}
\end{figure}

\noindent  but also recent methods such as DiffUIR, GSAD, UHDFormer and CIDNet. Under 256×256 input, U$^2$CLLIE maintains a lightweight design (0.48M params, 6.18 GFLOPs), significantly surpassing heavier models such as MIRNet (31.76M), SNR-Aware (39.12M), EnlightenGAN (114.35M), and CLE Diffusion (37.02M), striking a compelling balance between quality and efficiency.

\noindent \textbf{Qualitative Comparison. }Figure~\ref{vis_huawei} and~\ref{vis_real_synthetic} show visual compared results. Existing methods often fail to recover fine details or maintain color consistency in extremely dark regions, leading to noticeable noise and artifacts (\textit{e.g.}, gray/black shifts in the lamp area of Figure~\ref{vis_real_synthetic}). In contrast, U$^2$CLLIE benefits from entropy-guided feature extraction for robust representation in low-light regions, and asymmetric causal modeling that balances luminance enhancement with local structure and color fidelity.\par

%More results are included in the supplementary.

\noindent \textbf{Quantitative Evaluation on \textit{Unpaired} Datasets. }Table~\ref{tab_niqe} reports NIQE on five unpaired datasets. U$^2$CLLIE achieves the lowest (\textit{\textit{i.e.}}, best) scores across all cases, demonstrating superior perceptual quality. \par

\section{Ablation Study}

We conduct ablation studies on the LSRW-Huawei dataset under the same settings as \textit{"Implementation Details"}.

\noindent \textbf{Module Component. }Based on the results in Table~\ref{tab_ablation_loss}, the baseline with residual blocks yields 20.15 dB PSNR, 0.627 SSIM, and 0.177 LPIPS. Adding \textit{LEN} improves PSNR to 20.26 dB, showing the benefit of luminance-aware enhancement. \textit{NeCo} further raises PSNR to 21.14 dB and SSIM to 0.637, highlighting the effectiveness of neighborhood structural modeling. \textit{UaD} brings a 0.72 dB gain over the baseline by enhancing perceptual feature quality via uncertainty guidance. Removing \textit{AsC} or \textit{UaD}-\textit{AsC} from the full model causes clear performance drops, especially without \textit{AsC}. The complete U$^2$CLLIE achieves the best PSNR (21.35 dB), SSIM (0.633) and LPIPS (0,162), confirming the joint benefit of entropy-guided and causal-aware design. \par

\noindent \textbf{Loss Function. }
As shown in Table~\ref{tab_ablation_loss}, the Baseline ($\mathcal{L}_{mse} + \mathcal{L}_{ssim}$) yields unsatisfactory perceptual quality (PSNR: 20.85 dB, SSIM: 0.628, LPIPS: 0.173). Introducing L1 and L2, which incorporate feature-level consistency (Bseline + $\mathcal{L}_{per}$) and luminance distribution alignment (L1 + $\mathcal{L}_{global}$), significantly improves visual performance. L3 adds local color consistency($\mathcal{L}_{color}$), yielding the best LPIPS (0.161). The full model achieves the highest PSNR (21.35 dB) and SSIM (0.633), and the second-best LPIPS (0.162), balancing perceptual quality and reconstruction accuracy.
\setcounter{table}{2}
\begin{table}[H]
  \centering
  \setlength{\tabcolsep}{1mm}
  \begin{tabular}{lcccccc}
    \toprule
    \makecell[l]{Method} & 
    \makecell{LIME} & 
    \makecell{VV} & 
    \makecell{DICM} & 
    \makecell{NPE} & 
    \makecell{MEF} & 
    \makecell{AVG} \\
    \midrule
    \makecell[l]{Kind}           & 4.77 & 3.84 & 3.61 & 4.18 & 4.82 & 4.24 \\
    
    \makecell[l]{MIRNet}         & 6.45 & 4.74 & 4.04 & 5.24 & 5.50 & 5.19 \\ 
    \makecell[l]{SGM}            & 5.45 & 4.88 & 4.73 & 5.21 & 5.75 & 5.21 \\ 
    \makecell[l]{FECNet}         & 6.04 & 3.35 & 4.14 & 4.50 & 4.71 & 4.55 \\ 
    \makecell[l]{HDMNet}         & 6.40 & 4.46 & 4.77 & 5.11 & 5.99 & 5.35 \\ 
    \makecell[l]{Bread}          & 4.72 & 3.30 & 4.18 & 4.16 & 5.37 & 4.35 \\ \midrule
    \makecell[l]{Retinexformer}      & 3.44 & 3.71 & 4.01 & 3.89 & \underline{3.73} & 3.76 \\ \midrule
    \makecell[l]{FourLLIE}       & 4.40 & \underline{3.17} & \textbf{3.37} & 3.91 & 4.36 & 3.84 \\ \midrule
    \makecell[l]{DMFourLLIE}   & \underline{3.23} & 3.30 & 3.61 & \underline{3.56} & \underline{}{3.57} & \underline{3.46} \\ \midrule
    \makecell[l]{CWNet}          & 3.58 & 3.74 & 4.50 & 4.54 & 4.76 & 3.70 \\  \midrule
    \makecell[l]{\textbf{Ours}}  & \textbf{3.19} & \textbf{2.97} & \underline{3.48} & \textbf{3.31} & \textbf{3.18} & \textbf{3.23} \\ \midrule
    \bottomrule
  \end{tabular}
  \caption{NIQE scores (↓) of state-of-the-art methods on five unpaired datasets. The \textbf{best} and \underline{second-best} results are shown in \textbf{bold} and \underline{underlined}, respectively. ‘AVG’ denotes the average NIQE score over all five datasets. (All methods pre-trained on LSRW-Huawei.)}
  \label{tab_niqe}
\end{table}
\begin{table}[H]
  \centering
  \setlength{\tabcolsep}{1mm}
    \begin{tabular}{c|c|c|c}
    \toprule
    \multirow{2}[3]{*}{Method} & \multicolumn{3}{c}{Metrics} \\
\cmidrule{2-4}          & PSNR↑ & SSIM↑ & \multicolumn{1}{c}{LPIPS↓} \\
    \midrule
    \multicolumn{4}{c}{\textit{Module Component}} \\
    \midrule
    \makecell[c]{Baseline} & 20.15 & 0.627 & 0.177  \\
    \makecell[c]{Ours w/ \textit{LEN} Only} & 20.26 & 0.626 & 0.173  \\
     \makecell[c]{Ours w/ \textit{NeCo} Only} & 21.14 & 0.637 & 0.166  \\
     \makecell[c]{Ours w/ \textit{UaD} Only} & 20.87 & 0.630 & 0.170  \\
     \makecell[c]{Ours w/ \textit{AsC} Only} & 20.28 & 0.623  & 0.172  \\
     \makecell[c]{Ours w/o \textit{AsC}\&\textit{UaD}} & 21.07 & 0.631 & 0.164  \\
     \makecell[c]{Ours w/o \textit{AsC}} & 21.11 & 0.632 & 0.171  \\
    \midrule
    \multicolumn{4}{c}{\textit{Loss Function}} \\
    \midrule
     \makecell[c]{Baseline: ($\mathcal{L}_{mse}$+$\mathcal{L}_{ssim}$)} & 20.85 & 0.628 & 0.173  \\ \midrule
     \makecell[c]{L1: Baseline $\& \mathcal{L}_{per}$}  & 21.00  & 0.630 & 0.167  \\ \midrule
     \makecell[c]{L2: L1 $\& \mathcal{L}_{global}$}   & 20.97 & 0.631 & 0.166  \\ \midrule
    \makecell[c]{L3: L2 $\& \mathcal{L}_{color}$} & 21.05 & 0.631  & \textbf{0.161}  \\
    \midrule
    \midrule
     \makecell[c]{Ours} & \textbf{21.35} & \textbf{0.633} & 0.162  \\
    \bottomrule
    \end{tabular}%
     \caption{Ablation study on U$^2$CLLIE. Effectiveness of core modules and Loss components.}
  \label{tab_ablation_loss}%
\end{table}%
\noindent \textbf{Module Replacement. }To validate the architectural superiority of U$^2$CLLIE, we perform controlled replacements of key components with those from representative methods. As shown in Table~\ref{tab_ablation_replace}, replacing \textit{LEN} with the illumination branches from Retinexformer and CIDNet reduces PSNR to 20.77 dB and 20.98 dB, respectively. While CIDNet outperforms Retinexformer by 0.21 dB due to its HVI-aware spatial modeling, our design—using only a simple residual structure—achieves the best PSNR, SSIM and LPIPS scores, confirming that U$^2$CLLIE attains strong performance without architectural complexity. \par

% \noindent \textbf{Uncertainty Entropy Dominated Domain. }To assess the effectiveness of uncertainty-guided mechanisms under frequency-domain dominated, we design four experimental variants. As shown in Table 5 and Figure~\ref{ablation_vis_untertainty}, frequency-domain guidance consistently outperforms its time-domain counterparts. For instance, DMFourLLIE improves SSIM and LPIPS by capturing richer semantics and structured entropy distributions (Figure~\ref{ablation_vis_untertainty}, third column). Wavelet-domain guidance further enhances results, achieving the best SSIM (0.6391) and LPIPS (0.1607) with improved color consistency and reduced artifacts (Figure~\ref{ablation_vis_untertainty}, fourth column). Our full model, equipped with \textit{G2AF}-based frequency-domain guidance and adaptive Gaussian filtering, attains the highest PSNR (21.35 dB), demonstrating superior entropy stability, detail preservation, and color fidelity (Figure~\ref{ablation_vis_untertainty}, last column). 

\begin{table}[H]
  \centering
  \setlength{\tabcolsep}{1mm}
  \begin{tabular}{c|c|c|c}
    \toprule
    \multirow{2}[3]{*}{Ablation Settings} & \multicolumn{3}{c}{Metrics} \\
    \cmidrule{2-4}
     & PSNR↑ & SSIM↑ & LPIPS↓ \\
    \midrule
    \multicolumn{4}{c}{\textit{Module Replacement}} \\
    \midrule
    \textit{LEN} → Retinexformer & 20.77 & 0.629 & 0.164 \\ \midrule
    \textit{LEN} → CIDNet & 20.98 & 0.630 & 0.171 \\
    \midrule
    \multicolumn{4}{c}{\textit{Uncertainty Entropy Dominated Domain}} \\
    \midrule
    w/ Spatial Domain & 21.21 & 0.628 & 0.164 \\ \midrule
    \makecell[c]{w/ Frequency Domain\\\textit{G2AF} → DMFourLLIE} & 21.30 & 0.629 & 0.163 \\ \midrule
    \makecell[c]{w/ Frequency Domain\\\textit{G2AF} → WaveMamba} & 21.24 & 0.631 & 0.163 \\
    \midrule
    \midrule
    Ours & \textbf{21.35} & \textbf{0.633} & \textbf{0.162} \\
    \bottomrule
  \end{tabular}
  \caption{Comprehensive Ablation studies on U$^2$CLLIE framework design.}
  \label{tab_ablation_replace}
\end{table}

\noindent \textbf{Uncertainty Entropy Dominated Domain. }To evaluate the effectiveness of frequency-domain uncertainty guidance, we design four variants. As shown in Table~\ref{tab_ablation_replace}, frequency-domain methods consistently outperform spatial-domain ones. Replacing our G2AF with other frequency modules leads to PSNR, SSIM, and LPIPS drops, color distortion, and loss of detail (see Figure~\ref{ablation_vis_untertainty} for more details). \par

\textbf{Additional ablations and U$^2$CLLIE's generalization to downstream tasks are in the supplementary.}

\setcounter{figure}{4}
\begin{figure}[H]
    \centering
    \includegraphics[width=1.\linewidth]{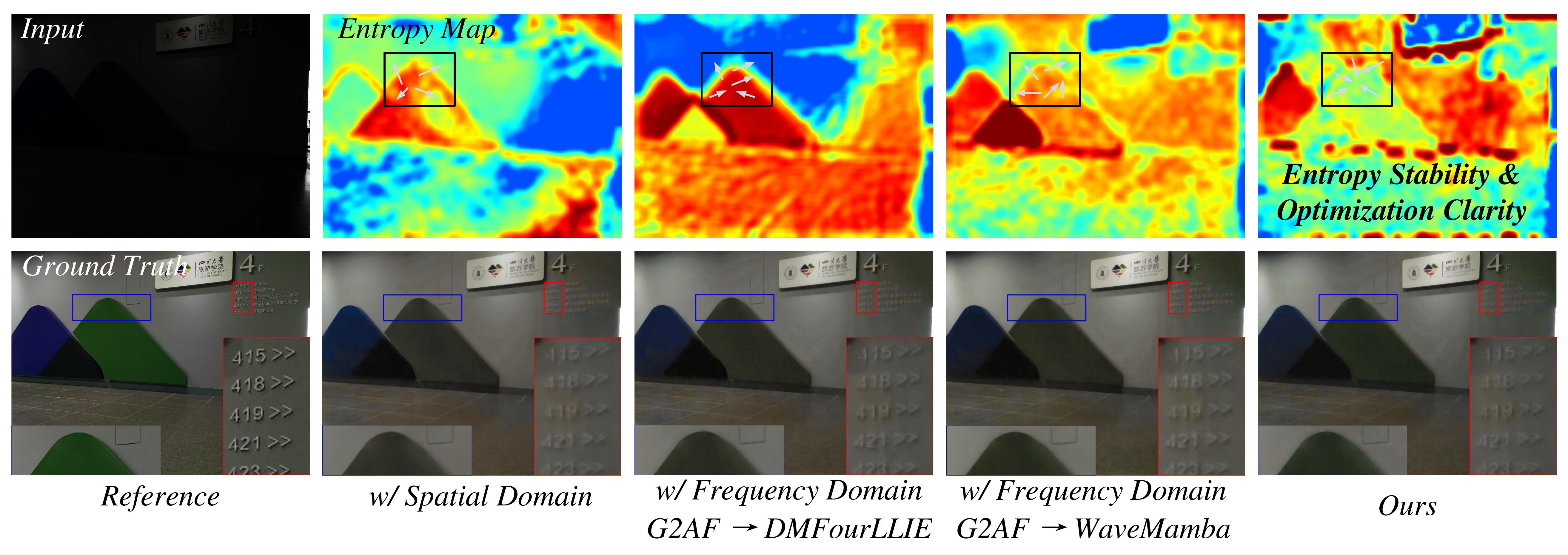}
    \caption{Uncertainty entropy distribution and visualization of color/detail reconstruction. (Arrow direction: high entropy → low entropy). U$^2$CLLIE exhibits stable entropy, clear optimization direction, consistent color (artifact-free), and enhanced detail (text clearly visible).~\textit{Please zoom in to the best view.}}
    \label{ablation_vis_untertainty}
\end{figure}

\section{Conclusion}

This paper presents U$^2$CLLIE, a novel low-light enhancement framework guided by frequency-domain uncertainty entropy. The \textit{UaD} module quantifies feature entropy to regulate gradients, alleviating noise dominance and optimization disorder in dark regions. Asymmetric causal modeling (\textit{NeCo} and \textit{AsC}) restores local consistency via neighborhood reasoning. Extensive experiments demonstrate that U$^2$CLLIE surpasses state-of-the-art methods in both metrics and perceptual quality, validating the effectiveness of entropy-guided enhancement.\par

\bibliography{aaai2026,references}

\end{document}